\documentclass{article}

\PassOptionsToPackage{numbers, compress}{natbib}

\usepackage[preprint]{arxiv}

\newtheorem{theorem}{Theorem}
\newtheorem{lemma}[theorem]{Lemma} 
\newtheorem{proposition}[theorem]{Proposition} 

\newtheorem{corollary}[theorem]{Corollary}
\newtheorem{definition}[theorem]{Definition}

\newtheorem{assumption}[theorem]{Assumption}

\usepackage[utf8]{inputenc} 
\usepackage[T1]{fontenc}    
\usepackage{hyperref}       
\usepackage{url}            
\usepackage{booktabs}       
\usepackage{amsfonts}       
\usepackage{nicefrac}       
\usepackage{microtype}      
\usepackage{xcolor}         
\usepackage{amsmath}
\usepackage{amssymb}            
\usepackage{mathtools}  
\usepackage{booktabs}

\usepackage{color}
\usepackage{xcolor}
\definecolor{blue}{rgb}{0,0.2,0.5}
\definecolor{green}{rgb}{0.1,0.35,0.0}
\definecolor{red}{rgb}{0.5,0.0,0.0}
\definecolor{purple}{rgb}{0.4,0,0.6}
\definecolor{cyan}{rgb}{0.0,0.4,0.3}
\definecolor{orange}{rgb}{0.6,0.4,0.0}
\definecolor{gray}{rgb}{0.3,0.3,0.3}

\hypersetup{%
  colorlinks,
  linkcolor=red,
  citecolor=blue, 
  urlcolor=blue   
}

\DeclareMathOperator*{\argmax}{argmax}

\title{Inverse Concave-Utility Reinforcement Learning \\ \emph{is} Inverse Game Theory}

\author{%
  Mustafa Mert Çelikok \\
  Delft University of Technology\\
  The Netherlands \\
  \texttt{m.m.celikok@tudelft.nl} \\
   \And
   Jan-Willem van de Meent \\
   University of Amsterdam \\
  The Netherlands \\
 \texttt{j.w.vandemeent@uva.nl} \\
    \AND
   Frans A. Oliehoek \\
   Delft University of Technology \\
   The Netherlands \\
\texttt{f.a.oliehoek@tudelft.nl} \\
}

\begin{document}

\maketitle

\begin{abstract}
We consider inverse reinforcement learning problems with concave utilities. Concave Utility Reinforcement Learning (CURL) is a generalisation of the standard RL objective, which employs a concave function of the state occupancy measure, rather than a linear function. CURL has garnered recent attention for its ability to represent instances of many important applications including the standard RL such as imitation learning, pure exploration, constrained MDPs, offline RL, human-regularized RL, and others. Inverse reinforcement learning is a powerful paradigm that focuses on recovering an unknown reward function that can rationalize the observed behaviour of an agent. There has been recent theoretical advances in inverse RL where the problem is formulated as identifying the set of feasible reward functions. However, inverse RL for CURL problems has not been considered previously. In this paper we show that most of the standard IRL results do not apply to CURL in general, since CURL invalidates the classical Bellman equations. This calls for a new theoretical framework for the inverse CURL problem. Using a recent equivalence result between CURL and Mean-field Games, we propose a new definition for the feasible rewards for I-CURL by proving that this problem is equivalent to an inverse game theory problem in a subclass of mean-field games. We outline future directions and applications in human--AI collaboration enabled by our results.
\end{abstract}

\section{Introduction}
\label{sec:intro}
 In this work, we present the first theoretical results and formalization of \emph{inverse concave-utility reinforcement learning} (I-CURL), which is the problem of rationalising an optimal CURL policy by inferring its reward function. In standard reinforcement learning, the objective is one way or another an expectation of return. It is an age-old result that this objective can be expressed as an inner-product between the reward function $R$ and the \emph{state-action occupancy measure} $d^\pi$ induced by a policy. However, in recent years, different applications and problems related to RL with objectives that break the linearity have emerged. As a result of this, concave utility reinforcement learning (CURL), also called convex RL, has garnered significant recent interest due to its ability to encompass a large variety of problems related to reinforcement learning (see table \ref{tbl:table} for a non-exhaustive list) \cite{zhang2020variational,10.5555/3535850.3535906, hazan2019provably, agarwal2022concave, mutti2023convex, zahavy2021reward, 10.5555/3635637.3663281}. However, the question of inverse RL for CURL problems has not been addressed before.
\begin{table}
\caption{Problems related to RL and their corresponding concave utility functions $F.$}
\begin{tabular}{@{} *5l @{}}    \toprule
\emph{Application} & \emph{The Objective $F$} &&&  \\\midrule
Standard RL    & $\langle d^\pi, R \rangle$  \\ 
Pure Exploration \cite{hazan2019provably} & $d^\pi \log d^\pi$  \\ 
Constrained MDPs \cite{altman2021constrained, borkar2005actor, tessler2018reward, calian2020balancing} & $\langle d^\pi, R \rangle \quad \text{s.t.} \quad \langle d^\pi, C \rangle \leq c$ \\ 
Offline RL \cite{10.5555/3535850.3535906} & $\langle d^\pi, R \rangle - \lambda KL(d^\pi, d^{\text{data}})$
\\ 
Imitation Learning \cite{ho2016generative, lee2019efficient, ghasemipour2020divergence} & $||d^\pi - d^E||^2_2, \quad KL(d^\pi, d^E)$  \\ 
Risk-sensitive RL \cite{tamar2015policy, chow2015risk, chow2018risk} & $\text{CVaR}_\alpha(\langle d^\pi, R \rangle), \quad \langle d^\pi, R \rangle -  \text{Var}(\langle d^\pi, R \rangle)$  \\
Cautious / Human-regularised RL \cite{Zhang2020CautiousRL, cornelisse2024human, jacob2022modeling} & $\langle d^\pi, R \rangle - \lambda KL(d^\pi, d^\rho)$, $d^\rho$ is a prior. \\ 
\bottomrule
 \hline

 \label{tbl:table}
\end{tabular}
\end{table}

\emph{Inverse reinforcement learning} (IRL, \cite{ng2000algorithms}) methods infer the reward function an agent is optimising for, under the assumption that the agent is following the standard RL objective of maximising the expected return. Consider observing an agent (AI or human) performing a task in the environment. We do not know what is their goal, however we would like to learn how to perform this task ourselves. We can try to clone the agent's behaviour directly, but a stronger objective is to \emph{rationalise} the behaviour of the agent by inferring their reward function. This is due to the fact that the reward function is a much more succinct and transferable definition of the task. 

Unfortunately, there are difficulties of applying IRL to infer reward functions from human demonstrations, due to bounded rationality \cite{simon1957models}. Humans have cognitive constraints that lead to deviations from perfectly rational behaviour. This is often modelled under a computational rationality framework, where a human's behaviour is optimal with respect to their cognitive constraints at the time \cite{lewis2014computational, gershman2015computational}. An established result in IRL presents a no-free-lunch theorem for IRL with humans, where trying to infer a feasible reward function without taking into account the humans' bounded rationality is in general not possible \cite{armstrong2018occam}. However, various forms of bounded rational behaviour can be represented as solutions to CURL problems, which is an important motivation for inverse CURL.

\begin{figure}
    \centering
\includegraphics[width=0.60\linewidth]{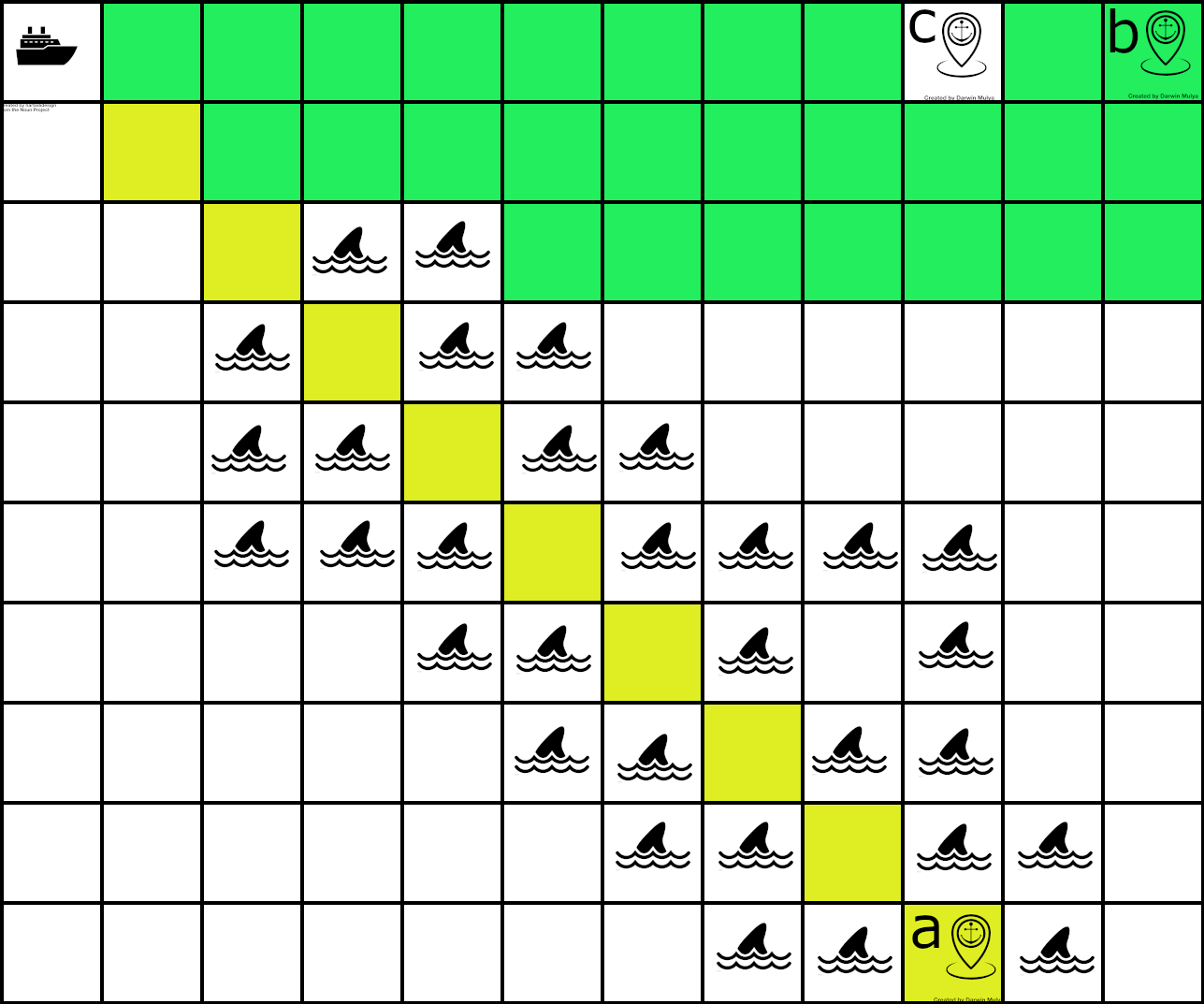}
    \caption{Illustrative example of an information-limited MDP.  A captain must navigate to one of the ports (a, b, c). The sharks denote dangerous waters with highly negative reward. The most preferred port is $a$ and the least preferred is $c$. The environment is not stochastic, but a cognitively constrained captain can make mistakes. The captain chooses to go to port B, knowing they could make  mistakes when following the golden path to a. The green-shaded area shows the paths captain follow to b. They make a special effort to avoid c, but otherwise follow a noisy path.} 
    \label{fig:maritime}
\end{figure}

Imagine that we are trying to infer the reward function of a human who has information-processing bounds, which is a form of bounded rationality due to cognitive constraints \cite{ortega2013information, ortega2013thermodynamics, e16084662,ho2020efficiency}. This setting is often modelled as information-limited decision-making that is equivalent to the cautious / human-regularised RL formulation in table \ref{tbl:table} \cite{rubin2012trading, genewein2015bounded, lancia2023humans}. Here, $d^\rho$ represents a prior, low mental-effort policy for the human, and they pay a cost for deviating from it proportional to the KL divergence. The figure \ref{fig:maritime} presents an illustrative example, where a captain must navigate to a port \footnote{The shark, ship, and port icons created by Gilberto, hartadidesign, and Darwin Mulya from the Noun Project.} . The captain's true reward function is such that each time-step has a small cost, crossing into dangerous waters (represented as sharks) have a high cost, and the rewards at the ports induce a preference ordering so that the port A is preferred the most, and C the least. The environment is not stochastic, thus actions are always executed correctly. Here, the obviously optimal policy for the captain is to deterministically follow the golden path. However, being under a heavy cognitive burden, the captain knows they are prone to making errors.  Thus instead, they follow a more forgiving path to port B, which leaves room for error. In similar episodes, they do make mistakes, and we end up observing a dataset of demonstrations within the green-shaded region. This is not the same as having a stochastic environment, since another captain might successfully execute the golden path in the same environment, and we already know the dynamics are deterministic. Here, even in the limit of infinite demonstrations, we cannot infer the true reward function with standard IRL. 

In this work, we address the question of \emph{how to formulate the problem of rationalising the behaviour of an agent whose policy is solving a CURL problem.} In section \ref{sec:curl}, we start by presenting the formal definition of CURL and an important result from literature that shows how it cannot be reduced to a standard RL problem in general. Then in section \ref{sec:mfg}, we explain how every CURL problem is proven to be equivalent to a mean-field game (MFG). A core contribution of our work is taking advantage of this result to prove that IRL in CURL problems is in fact equivalent to inverse game theory (IGT) in mean-field games. In order to make this connection clear, we give a brief background of the standard IRL and inverse game theory in sections \ref{sec:irl} and \ref{sec:igt}. 

In section \ref{sec:i-curl}, we prove how the feasible reward set definition of standard IRL fails in inverse CURL (I-CURL). Then using the equivalence to MFGs, we propose a new set, and prove that it is sufficient and necessary, containing all rewards that rationalise the optimal CURL policy. We show an equivalent characterization of this set where computing a feasible reward function is solving a constant-sum game itself. Since the inverse CURL problem has not been addressed before, our work fills an important theoretical gap, demonstrating the limitations of standard IRL methods for I-CURL. Our formulation enables the inference of reward functions from bounded rational behaviour such as information-boundedness and risk-aversion.

\section{Background}
\label{sec:background}
\subsection{Concave Utility Reinforcement Learning}
\label{sec:curl}
\begin{definition}[Markov Decision Process]
    A Markov decision process (MDP) is defined by the tuple $M= (\mathcal{S}, \mathcal{A}, T, R, \mu_0)$ where the $\mathcal{S}$ and $\mathcal{A}$ are finite sets of states and actions, the $T: S \times A \rightarrow \Delta(S)$ is the transition kernel that maps state-action pairs to distributions over states, $R: S \times A \rightarrow [0,1]$ is the reward function, and $\mu_0 \in \Delta(S)$ is the initial state distribution. 
\end{definition}

When paired with a memoryless policy $\pi: \mathcal{S} \rightarrow \Delta(\mathcal{A})$ that maps states to distributions over actions, the MDP $M$ induces a probability measure $P^\pi(\tau)$ over infinite-length trajectories $\tau=(S_0, A_0,....)$  where $P^\pi(S_0=s) = \mu_0(s),$ $P^\pi(A_t= a | S_0,A_0,...,S_t) = \pi(a | S_t),$ and $P^\pi(S_{t+1} = s | S_t, A_t) = T(s | S_t, A_t).$
An MDP is usually coupled with an optimality criterion $OPT$, and unless stated otherwise, we assume the infinite-horizon expected sum of discounted rewards criterion, $OPT^\infty(\pi) \triangleq \mathbb{E}_{P^\pi}[\sum^\infty_{t=0} \gamma^t R(s_t, a_t)].$ Essentially, an optimality criterion induces a partial ordering amongst policies and provides us with the criterion we use for deciding how good a policy is. The optimal policy is then defined as $\pi^* \in \argmax_\pi OPT^{\infty}(\pi).$ 
\begin{definition}[Discounted State-Action Occupancy Measure]
  When connected to an MDP $M$, every policy $\pi$ induces a \emph{discounted state-action occupancy measure} defined as $d^\pi(s,a) \triangleq (1-\gamma) \sum^\infty_{t=0} \gamma^t Pr(S_t = s, A_t = a | S_0 \sim \mu_0, \pi).$  
\end{definition}
In essence $d^\pi(s,a)$ tells us the joint probability of visiting state $s$ and taking action $a$, under policy $\pi.$ From hereon, we will let $d^\pi$ denote the measure induced by the policy $\pi$ in an MDP that will be clear from the context. The set of all possible discounted state-action occupancy measures forms a polytope that is a convex and compact set (\cite{puterman2014markov}) and it is defined as follows. 
\begin{definition}[Set of Discounted State-Action Occupancy Measures]
    For a given MDP $M$ and discount factor $\gamma$, the set of all possible discounted state-action occupancy measures is defined as $\mathcal{K}_\gamma = \{ d^\pi | d^\pi \geq 0, \sum_{a \in A} d^\pi(s,a) = (1-\gamma) \mu_0(s) + \gamma \sum_{s', a'}T(s| s',a')d^\pi(s',a') \quad \forall s \in S\}.$
\end{definition}
A standard result in RL is that the $OPT^\infty(\pi)$ can equivalently be expressed as a linear function of $d^\pi$, $\langle d^\pi, R \rangle$. In that case, the standard reinforcement learning objective of finding the optimal policy can be cast as the following linear program 
\begin{equation}
\max_{d^\pi \in \mathcal{K}_\gamma} \langle d^\pi, R \rangle. \qquad \textbf{(Standard RL Objective)}
\label{eqn:standard-RL}
\end{equation} 
\emph{Concave utility reinforcement learning (CURL)} replaces the linear objective function of equation \ref{eqn:standard-RL} with a concave function $F: \Delta(\mathcal{S} \times \mathcal{A}) \rightarrow (-\infty, K]$ of the state-action occupancy measures. Then the new optimization problem becomes the following convex program 
\begin{equation}
    \max_{d^\pi \in \mathcal{K}_\gamma} F(d^\pi). \qquad \textbf{(CURL Objective)}
\end{equation} 
Solving CURL problems is not as straightforward as applying standard RL methods directly, demonstrated by the following lemma from \citet{zahavy2021reward}.
\begin{lemma}
\label{thm:curl-reward-lemma}
        There exists an MDP $M$ with concave utility $F$ such that there can be no stationary reward $R \in \mathbb{R}^{S \times A}$ with  $\argmax_{d^\pi \in \mathcal{K}_\gamma} \langle d^\pi, R \rangle = \argmax_{d^\pi \in \mathcal{K}_\gamma} F(d^\pi).$
\end{lemma}
The lemma above shows that in general, CURL problems cannot be reduced into standard RL problems by re-defining rewards. In other words, we cannot be sure of an equivalent stationary reward function that only depends on $(s,a).$ In addition, unlike in standard RL, a deterministic optimal policy may not exist for a CURL problem \cite{10.5555/3535850.3535906,zahavy2021reward,mutti2023convex}.

\subsection{CURL as Mean-field Games}
\label{sec:mfg}
In a bid to develop methods that can solve CURL problems, two concurrent formulations have emerged: CURL as mean-field games \cite{10.5555/3535850.3535906} and CURL as zero-sum games \cite{zahavy2021reward}. As we show in appendix \ref{apn:curl-zerosum}, these two formulations are strongly related. Here, we will focus on the mean-field games.

Any CURL problem can be formulated as a special case of a mean-field game (that belongs to a sub-class of MFGs) \cite{10.5555/3535850.3535906}. In this sub-class, an MFG is a tuple $G=(S,A,T, R, \gamma, \mu_0)$ where the only difference from MDPs is in the reward function\footnote{We overload the reward notation $R$. Unless stated otherwise, $R$ will be of form $R(s,a,d).$} $R: S \times A \times \Delta(\mathcal{S} \times \mathcal{A}) \rightarrow \mathbb{R}.$ The interpretation is that, in a large population of agents, $R(s,a,d)$ represents the reward an agent receives at state $s$ by playing action $a$, while the population policy induces the distribution $d$. For a fixed mean-field distribution $d$, we can write the objective of an agent in the MFG as $\langle d^\pi, R(\cdot, \cdot, d) \rangle,$ which is the expected sum of rewards the agent would receive by playing policy $\pi$ while the rest of the population continues inducing $d.$ The core assumption in MFGs is the fact that the changes in one agent's policy has no influence on the mean-field distribution. Theorem 1 of \citet{10.5555/3535850.3535906} proves that every CURL problem is in fact such an MFG with the reward $R(\cdot, \cdot, d^\pi) = \nabla F(d^\pi) \in \mathbb{R}^{S \times A}$ \footnote{The $d^\pi$ and thus $\nabla F(d^\pi)$ can be flattened into a vector of length $|\mathcal{S} \times \mathcal{A}|$ when necessary.}, where any solution to the CURL problem is a mean-field Nash equilibrium (definition \ref{defn:mfne}) and vice versa. From now on, we will refer to MFGs induced by a CURL problem as CURL-MFGs.

\begin{definition}[Mean-field Nash Equilibrium (MFNE) and Exploitability]
    Define the exploitability of policy $\pi$ in a CURL-MFG as $\phi(\pi) = \max_d \langle d, R(\cdot, \cdot, d^\pi) \rangle - \langle d^\pi, R(\cdot, \cdot, d^\pi) \rangle$. An MFNE (or equivalently the optimal policy for the CURL problem) is a tuple $(\pi^* , d^{\pi^*})$ such that $\phi(\pi^*) = 0$. 
    \label{defn:mfne}
\end{definition}

\subsection{Inverse Reinforcement Learning}
\label{sec:irl}
Let $\Bar{M}=(S, A, T, \mu_0)$ denote an MDP without reward. An IRL problem is formalized by the tuple $\mathbb{B}=(\Bar{M}, \pi^E)$ where $\pi^E$ is an expert policy that behaves optimally in $\Bar{M}$ with regards to an unknown reward function. The standard IRL problem has emerged as rationalising $\pi^E$ by inferring the reward function it is optimal for. However, this problem is ill-conditioned since in general the reward function is not unique. Recent works in IRL have then re-defined the problem as a well-posed one \cite{metelli2021provably, metelli2023towards, lindner2022active}: inferring the set of all reward functions compatible with $\pi^E$, denoted by $\mathcal{R}_{\mathbb{B}}.$ As defined below, the set condition requires non-positive advantage everywhere, since a positive advantage value would mean there is at least one state where deviating from $\pi^E$ improves return.
\begin{definition}[Feasible Rewards Set for IRL \cite{metelli2021provably, metelli2023towards, lindner2022active}]
 The set of feasible rewards for the IRL problem $\mathbb{B}$ is defined as $\mathcal{R}_{\mathbb{B}} = \{ R \in \mathbb{R}^{S \times A} | A^{\Bar{M} \cup R}_{\pi^E} \leq 0\},$ where $A^{\Bar{M} \cup R}_{\pi^E}(s,a) = Q^{\Bar{M} \cup R}_{\pi^E}(s,a) - V^{\Bar{M} \cup R}_{\pi^E}(s)$ is the advantage function of $\pi^E$ computed according to $R.$
 \label{defn:standard-reward-set}
\end{definition}

Unfortunately in practice, we do not have access to $\pi^E$, but have a dataset of state-action pairs generated by $\pi^E$ \footnote{We assume $T$ is known, but it also can be estimated from data}. When the expert policy is estimated from a dataset as $\hat{\pi}^E$, it induces an \emph{empirical} IRL problem denoted as $\hat{\mathbb{B}} = (\Bar{M}, \hat{\pi}^E)$ with its feasible reward set $\mathcal{R}_{\hat{\mathbb{B}}}$ defined accordingly. One of the main questions studied by this line of research is how close $\mathcal{R}_{\hat{\mathbb{B}}}$ is to the true set of feasible rewards $\mathcal{R}_{\mathbb{B}}$. Under certain assumptions, it is possible to derive sample complexity bounds with respect to this gap within a PAC framework \cite{metelli2021provably, metelli2023towards, lindner2022active}. 

\subsection{Inverse Game Theory}
\label{sec:igt}
Inverse game theory (\cite{kuleshov2015inverse}) essentially deals with the problem of inferring reward functions of agents from observed equilibrium behaviour in games. In other words, IGT asks the question: "Which reward functions would rationalize the observed behaviour in a game?" Even though the IGT as a term is relatively recent, it has a long history in economics and optimization where it is called different names such as model estimation, inverse optimization, and computational rationalization. IRL and IGT share common history and methodology such as the maximum entropy methods \cite{waugh2011computational}. Since we are interested in formulating I-CURL problems within IGT, we will focus on IGT in mean-field games. 

\paragraph{IGT for Mean-field Games.}
There are two main approaches to IGT for MFGs: the population-level and the individual-level. The former focuses on learning a reward function of the form $R(d^\pi, \pi)$ whereas the latter focuses on learning the individual agent's reward $R(s, a, d)$  from a dataset consisting of $(s_i,a_i) \sim d^*$ where $(\pi^*, d^*)$ is the MFNE. Therefore in this work, we will use the individual-level formulation presented in \citet{chen2022individual} as a starting point for the problem formulation of I-CURL. We let $\mathbb{B} = (\Bar{G}, \pi^E, d^E)$ denote an individual-level IGT problem where $\Bar{G}$ is an MFG without reward and $(\pi^E, d^E)$ is an MFNE of $\Bar{G}$ with respect to an unknown reward function.

\section{Inverse Game Theory for CURL}
\label{sec:i-curl}
Both inverse RL and inverse game theory share the same objective: to find a reward function that would rationalize the observed behaviour of agents. Since the lemma \ref{thm:curl-reward-lemma} states there cannot be a stationary reward function $R(s,a)$ in general for CURL problems, the standard formulation of IRL does not apply any more. In particular, the definition of the feasible rewards set $\mathcal{R}_{\mathbb{B}}$ is problematic for the CURL setting. It assumes that $\pi^E$ is optimising for the standard RL objective with stationary reward functions that only depend on $(s,a)$ and the set condition relies on the advantage function. However, as we know from the lemma $\ref{thm:curl-reward-lemma}$, there are optimal CURL policies that cannot be induced by such reward functions via the linear objective. Let $\mathbb{C} = (\Bar{M}, \pi^E)$ denote an I-CURL problem where $\pi^E$ is the optimal policy for a concave utility function $F$. The following lemma states that even in the exact case with true knowledge of $\pi^E$, the true reward function may be impossible to infer with standard IRL.
\begin{lemma}
    There exists an MDP $M$ with concave utility $F$ and reward function $r^*$ such that, given its I-CURL problem $\mathbb{C}= (\bar{M}, \pi^E)$, $r^* \notin \mathcal{R}_{\mathbb{C}}$ where $\mathcal{R}_{\mathbb{C}}$ is the feasible reward set for a standard IRL problem, as defined in definition \ref{defn:standard-reward-set}.
    \label{lemma:impossible-set}
\end{lemma}
\emph{Proof.} Consider the gridworld shown in \ref{fig:maritime} with rewards $r^*(\textit{port A}) = 10, r^*(\textit{port B}) = 5$, $r^*(\textit{port C}) = 1$, and $r^*(\textit{shark}) = -100.$ The reward is zero elsewhere. As described, the observed $\pi^E$ is the green-shaded region. The uniquely optimal policy under the standard RL objective is the deterministic golden path shown on the figure. Therefore, the advantage function of any other policy will not be $\leq 0$ everywhere. Thus, $r^*$ will not be in $\mathcal{R}_{\mathbb{C}}.$

\subsection{A Game-theoretic Feasible Reward Set for I-CURL}

The lemma \ref{lemma:impossible-set} motivates the need for a new definition for sets of feasible rewards in I-CURL problems. Let $\mathbb{B} = (\Bar{G}, \pi^E, d^{E})$ denote an individual-level IGT problem where $\Bar{G}$ is a CURL-MFG without reward and $(\pi^E, d^{E})$ is an MFNE in $\Bar{G}$ for some reward function $R.$ We propose the following definition as the feasible set of reward functions and prove that it is necessary and sufficient.
\begin{proposition}[Feasible Reward Set for I-CURL]
The set of feasible reward functions for the I-CURL problem formulated as the individual-level IGT in CURL-MFGs, $\mathbb{B} = (\Bar{G}, \pi^E, d^{E})$, is defined as $\mathcal{R}_{\mathbb{B}} = \{R \in \mathbb{R}^{S \times A \times \mathcal{K}_\gamma} | \phi(\pi^E; R) = 0 \}$ where the parameterized exploitability is $\phi(\pi^E; R) \triangleq  \max_d \langle d, R(\cdot, \cdot, d^{E}) \rangle - \langle d^{E}, R(\cdot, \cdot, d^{E}) \rangle $.
\label{prop:icurl-explicit}
\end{proposition}

\emph{Proof.} The feasible set condition simply means that the reward function $R$ must make $(\pi^E, d^{E})$ an MFNE. This is a sufficient and necessary condition, since the theorem 1 of \citet{10.5555/3535850.3535906} proves that any solution to the CURL problem is an MFNE in its CURL-MFG, and vice-versa. 

An important point to notice is the fact that even though $\mathcal{S}$ and $\mathcal{A}$ are finite, the CURL-MFG reward functions are not tabular since the $\mathcal{K}_\gamma$ is in general infinite. This means in practice we must choose a function class when solving I-CURL problems. Throughout this paper, we will make the following assumption.
\begin{assumption}[Convex parameterization]
    The CURL-MFG reward functions $R(\cdot, \cdot, d; \theta)$ are parameterized by $\theta \in \Theta$ where $\Theta$ is a non-empty, compact and convex set. Then the feasible reward set is expressed more succinctly as $\mathcal{R}_{\mathbb{B}} = \{\theta \in \Theta | \phi(\pi^E; \theta) = 0 \}.$
\end{assumption}

For a given $\theta$ and $\pi^E$, computing $\phi(\pi^E; \theta)$ involves two steps: solving a standard RL problem and performing policy evaluation. Specifically, computing the term $\max_d \langle d, R(\cdot, \cdot, d^{E}; \theta) \rangle$ is equivalent to solving the MDP $M^{\pi^E} = \Bar{M} \cup R^{\pi^E}$ where the $R^{\pi^E}(s,a) = R(s,a,d^{E};\theta)$. Then the second term is equivalent to evaluating $\pi^E$ in $M^{\pi^E}.$ 

\begin{proposition}
    The set $\mathcal{R}_{\mathbb{B}}$ is equivalent to the set of $\theta$ that solves the saddle-point problem $\min_{\theta \in \Theta}\{\max_{d \in K_\gamma} \langle d, R(\cdot, \cdot, d^{E};\theta)\rangle - \langle d^{E}, R(\cdot, \cdot, d^{E};\theta) \rangle\}.$
    \label{prop:saddle-point}
\end{proposition}
The proposition \ref{prop:icurl-explicit} can be seen as the explicit characterisation of the feasible reward set, whereas the saddle-point formulation of proposition \ref{prop:saddle-point} presents an alternative that is useful for theoretical analysis. This means we can formulate the problem of finding a feasible reward function as a min-max game itself with a parameterized utility function $U(\theta, d; d^{E}) = \langle d, R(\cdot, \cdot, d^{E};\theta) \rangle - \langle d^{E}, R(\cdot, \cdot, d^{E}; \theta) \rangle$ where $d^{E}$ is a parameter. This is in-line with recent concurrent work by \citet{goktas2024generative}. Unlike their case, we do not explicitly need to assume the maximisation is concave. We know that the set $\mathcal{K}_\gamma$ is a convex polytope, and therefore for a fixed $\theta$, the problem is concave in $d$, since $U$ is a linear function of $d$. Therefore we only need the following assumptions to extend their results to CURL-MFGs.
\begin{assumption}
  (Convex exploitability) The parameterized exploitability $\phi(\pi^E; \theta)$ of the expert policy is convex in $\theta$ and (Lipschitz-smoothness) the reward functions are $\ell$-Lipschitz smooth.
    \label{assm:convexity}
\end{assumption}
This makes the saddle-point formulation of proposition \ref{prop:saddle-point} a convex-concave game where $\ell$-Lipschitz smoothness allows for efficient solutions using gradient descent-ascent. Therefore the adversarial inverse multi-agent planning (AIMP) algorithm of \citet{goktas2024generative} can be directly applied. Then the complexity result of AIMP extends to our problem as follows.

\begin{corollary}
    Let $\theta_\epsilon$ be an $\epsilon$-feasible reward function such that $\phi(\pi^E; \theta_\epsilon) \leq \epsilon,$ or in other words, the $\pi^E$ is an $\epsilon$-MFNE for the reward function $R(\cdot, \cdot, d; \theta_\epsilon)$. Under the assumption \ref{assm:convexity}, the AIMP can compute a $\theta_\epsilon$ in O($\frac{1}{\epsilon^2})$ iterations.
    \label{thm:aimp}
\end{corollary}

These results highlight the difficulty of solving I-CURL problems exactly, even if the $\pi^E$ is known. This make intuitive sense, since the $\mathcal{K}_{\gamma}$ is infinite regardless of $\mathcal{S}$ and $\mathcal{A}$, and the feasible set condition is an equilibrium requirement. However, if the set of reward functions can be constrained to some well-behaving function classes, we can have a convex-concave game that can be solved by gradient descent-ascent methods as stated by the theorem above. Unfortunately, in real-world applications, we do not have access to $\pi^E$, but only a dataset of demonstrations from it. This induces the empirical I-CURL problem, defined and analysed as follows.

\subsection{Empirical I-CURL}
As in the IRL setting, we now define the empirical version of the individual-level IGT problem of a CURL-MFG $\mathbb{B} = (\bar{G}, \pi^E, d^{E}).$ Denoted by $\hat{\mathbb{B}}$, the empirical problem emerges when the expert policy $\pi^E$, and thus the MFNE $(\pi^E, d^{E})$ is not known. Here, the assumption is that we have a dataset $\mathcal{D}=\{(s_1, a_1),...,(s_N, a_N)\}$ of $N$ state-action pairs sampled from the equilibrium $(\pi^E, d^{E})$ where $(s_i, a_i) \sim d^E(s,a).$ The goal is to infer a reward function of the form $R(s,a,d)$ that rationalises the observed behaviour. There are two challenges here: (1) The empirical estimation of $(\pi^E, d^{E})$ from data and (2) The computation of $\mathcal{R}_{\hat{\mathbb{B}}}$. The $d^{E}$ is estimated with the empirical distribution $\hat{d}^{E}(s,a) = \frac{\sum^{N}_{i = 0}\mathbb{I}[(s_{i}, a_{i}) = (s,a)]}{N}$ if $(s,a)$ in the dataset, and $\frac{1}{SA}$ otherwise. Then the $\hat{\pi}^E(a | s) = \frac{\hat{d}^{E}(s,a)}{\hat{d}^{E}(s)}$ where $\hat{d}^{E}(s) = \sum_a \hat{d}^{E}(s,a).$ Then the 
empirical feasible set $\mathcal{R}_{\hat{\mathbb{B}}}$ is defined as 
\begin{equation}
    \mathcal{R}_{\hat{\mathbb{B}}} =  \{\theta \in \Theta | \phi(\hat{\pi}^E; \theta) = 0 \}.
\end{equation}

An important object to study is how $\mathcal{R}_{\mathbb{B}}$ relates to $\mathcal{R}_{\hat{\mathbb{B}}}$ in terms of the quality of the estimation $\hat{\pi}^E.$ It is clear that $\hat{\pi}^E \rightarrow \pi^E$ as the dataset grows so that $N \rightarrow \infty,$ and also self-evident that if $\hat{\pi}^E = \pi^E$ then $\mathcal{R}_{\hat{\mathbb{B}}} =\mathcal{R}_{\mathbb{B}}.$ Indeed the theorem 1 of \citet{chen2022individual} proves that in the asymptotic case we can recover at least some elements in $\mathcal{R}_{\mathbb{B}}.$ However, this does not say anything about the finite sample setting: even if $\hat{\pi}^E$ is close to the $\pi^E$, this does not give any guarantees on the similarity of the reward sets. The proposition \ref{prop:saddle-point} suggests that any such result would be bounding the difference between the solution sets of two different constant-sum games with utilities $U(\theta, d; d^{E})$ and $U(\theta, d; \hat{d}^{E}).$ With no further assumptions, proving any such result is difficult. 

\section{Related Works}

The recent work on inverse reinforcement learning has focused identifying the set of all feasible reward functions instead of a single reward function \cite{metelli2021provably, metelli2023towards, lindner2022active}. This re-formulation resolves the ill-posedness of the IRL problem and provides a more robust theoretical approach to analysing the complexity of IRL. In our work we follow the same approach for I-CURL problems. 

Inverse decision modelling was proposed by \citet{jarrett2021inverse} as an approach to learn more interpretable and expressive representations of bounded rational behaviour. I-CURL follows the same line of reasoning, since CURL can represent various behaviours due to cognitive constraints. An important line of work in recent years has been considering different forms of bounded rationality in inverse RL \cite{evans2016learning, majumdar2017risk, chan2021human, singh2018risk, shah2019feasibility}. This is due to an important result showing efficient reward inference is not possible without making assumptions about the agent's rationality \cite{armstrong2018occam}. Inverse RL has been a contender for solving the value alignment problem by inferring the constraints and goals of humans through inferring the reward function, eventually leading to AI agents that act according to the objectives of human users \cite{hadfield2016cooperative, malik2018efficient}. Recent work has applied IRL to the problem of aligning AI agents with human norms \cite{peschl2022moral}. 

The goal of population-level approaches is to infer the population-level reward of the form $R(d^\pi, \pi)$ from a dataset consisting of occupancy measure and policy pairs $(d_0, \pi_0),...,(d_T, \pi_T)$ instead of state-action tuples \cite{yang2018learning}. Our approach is based on  individual-level inverse game theory for mean-field games, which has been explored empirically for general mean-field games \cite{chen2022individual, chen2023adversarial}. The problem of rationalising observed multi-agent behaviour by inferring rewards has been referred to as inverse multi-agent RL \cite{natarajan2010multi}, inverse game theory \cite{kuleshov2015inverse}, and inverse equilibrium problems \cite{waugh2011computational} in different contexts. Recent formulation by \citet{goktas2024generative} represented the IGT problem itself as a zero-sum game whose solution set is referred to as the inverse Nash equilibria. 
\section{Discussion}


As discussed in Section \ref{sec:intro}, the bounded rationality of human behaviour can often be modelled as a CURL policy. For instance, the cautious reinforcement learning formulation by \citet{Zhang2020CautiousRL} represents an agent acting cautiously by staying close to their prior behaviour. Similarly, information-bounded decision-making models the cognitive cost of decision-making as a divergence from a low-effort baseline policy. Further human cognitive constraints can be framed within a constrained MDP, presenting a CURL problem. In this context, bounded rationality (also known as resource-rationality) implies that observed human behaviour reflects an optimal policy under these constraints. Consequently, our inverse CURL approach is crucial for learning reward functions and preferences from human behaviour. This argument is supported by recent works advocating for cognitive models of humans as inverse models in robotics and for making machine learning models more human-centric by incorporating user modeling pipelines \cite{ho2022cognitive, ccelikok2023modeling}. In multi-agent settings, where humans cooperate with and oversee AI agents, additional cognitive costs and constraints can significantly impact outcomes \cite{ccelikok2022best}. Furthermore, as noted by \citet{jarrett2021inverse}, an inverse RL methodology that accommodates bounded rational behaviour models leads to more interpretable reward functions and task descriptions. Therefore the ability to learn reward functions and task descriptions from bounded rational human behaviour is crucial for enhanced human–AI collaboration and alignment.

\paragraph{Limitations and Future Work.} In this work, we assumed known transition dynamics. Even though a common assumption in standard IRL, there are practical applications where this may not be realistic. Thus, relaxing this assumption is an important future direction. In this work, we focused on formulating the novel problem of I-CURL and theoretically analysing its feasibility. A direction this opens up is implementing the individual-level IGT for mean-field games and showing its empirical performance for I-CURL with different function classes. We have focused on the MFG formulation of CURL, and an interesting direction is whether a similar formulation can be derived for the zero-sum formulation of \citet{zahavy2021reward}. Finally, perhaps the most interesting direction is to apply I-CURL to learning interpretable reward functions from different models of bounded rational human behaviour such as information-boundedness or risk-aversion.

\section{Conclusion}
Standard inverse reinforcement learning is not directly applicable to inferring reward functions of agents that have optimized their policies for solving a concave utility reinforcement learning problem. Even in the idealized case of having access to the expert policy $\pi^E$, we have shown that the set of feasible reward functions as defined by the standard IRL literature can exclude the true reward function. We have resolved this problem by proving a game-theoretic characterization of the feasible reward set for inverse CURL, leveraging a result that shows every CURL problem is equivalent to a mean-field game.

\begin{ack}
This research was (partially) funded by the Hybrid Intelligence Center, a 10-year programme funded by the Dutch Ministry of Education, Culture and Science through the Netherlands Organisation for Scientific Research, https://hybrid-intelligence-centre.nl.
\end{ack}

\bibliography{Main/sample}
\bibliographystyle{unsrtnat}


\appendix

\section{Appendix / supplemental material}

\subsection{The Connection between Zero-sum and Mean-field Formulations of CURL}
\label{apn:curl-zerosum}

Any CURL problem can be re-formulated as a zero-sum game with convex-concave payoffs \cite{zahavy2021reward}. To explain, we need the following definition.
\begin{definition}[Fenchel Conjugate]
The Fenchel conjugate of a function $f$ is defined as $f^*(y) \triangleq \sup_x \langle x, y \rangle - f(x).$
\end{definition}
Fenchel conjugate is often called convex conjugate. Even though there is an equivalent definition of concave conjugate with $\inf$ instead of $\sup$, to go along with the established notation, we will use the convex equivalent of the CURL problem, $\min_{d^\pi \in \mathcal{K}_\gamma}E(d^\pi)$ where $E=-F.$ Let $E^*$ be the Fenchel conjugate of $E.$ Since $E$ is convex (and semi-continuous), the conjugate of its conjugate is equal to itself, $(E^*)^* = E.$ Then, the CURL problem can be re-written as the convex-concave zero-sum game 
\begin{equation}
    \min_{d^\pi \in \mathcal{K}_\gamma}E(d^\pi) = \min_{d^\pi \in \mathcal{K}_\gamma}\max_{\lambda \in \Lambda} \langle d^\pi, \lambda \rangle - E^*(\lambda),
\end{equation} 
where $\Lambda = \{ \nabla E(d^\pi) | d^\pi \in \mathcal{K}_\gamma\}$ is the closure of the (sub-)gradient space. An interesting point to notice here is the fact that for fixed $\lambda$, the minimisation problem can be thought of as a standard reinforcement learning problem with reward function $-\lambda.$ 
We can connect this zero-sum formulation to the mean-field one as follows. Since $E = - F$ and $F$ is concave, the $\nabla E$ is the same as $-\nabla F$. Then $\Lambda = \{ -\nabla F(d) | d \in \mathcal{K}_\gamma \} = \{ -R(\cdot, \cdot, d) | d \in \mathcal{K}_\gamma\}$ where $R(\cdot, \cdot, d)$ is the reward function of the CURL-MFG. Then the zero-sum game can be equivalently expressed as 
\begin{equation}
\min_{d^\pi \in \mathcal{K}_\gamma}\max_{d \in \mathcal{K}_\gamma} \langle d^\pi, -R(\cdot, \cdot, d) \rangle - E^*(-R(\cdot, \cdot, d)).    
\end{equation}


\end{document}